\documentclass{article}
\usepackage{spconf,amsmath,graphicx}

\usepackage{enumitem}
\usepackage{float}     
\usepackage{placeins}  
\usepackage{amsmath,amssymb,mathtools}
\usepackage{dsfont}

\setlist{nosep, leftmargin=14pt}

\usepackage{mwe} 
\usepackage{multirow} 
\newcommand{\Ind}{\mathds{1}}

\title{NullBUS: Multimodal Mixed-Supervision for Breast Ultrasound Segmentation via Nullable Global--Local Prompts}
\name{Raja Mallina*  \qquad\qquad\qquad\qquad\qquad\qquad   Bryar Shareef* }
\address{*Department of Computer Science, University of Nevada, Las Vegas, NV 89154, USA}
\begin{document}
%
\maketitle
\begin{abstract}
Breast ultrasound (BUS) segmentation provides lesion boundaries essential for computer-aided diagnosis and treatment planning. While promptable methods can improve segmentation performance and tumor delineation when text or spatial prompts are available, many public BUS datasets lack reliable metadata or reports, constraining training to small multimodal subsets and reducing robustness. We propose NullBUS, a multimodal mixed-supervision framework that learns from images with and without prompts in a single model. To handle missing text, we introduce nullable prompts, implemented as learnable null embeddings with presence masks, enabling fallback to image-only evidence when metadata are absent and the use of text when present. Evaluated on a unified pool of three public BUS datasets, NullBUS achieves a mean IoU of 0.8568 and a mean Dice of 0.9103, demonstrating state-of-the-art performance under mixed prompt availability.
\end{abstract}

\section{Introduction}

Breast cancer is the most commonly diagnosed cancer in U.S. women and a major cause of death, with an estimated 316{,}950 new invasive cases and 42{,}680 deaths in 2025 \cite{ACS_Breast_Common_2025}. Early detection and accurate lesion localization are essential for improving diagnostic outcomes and guiding treatment. Mammography remains the main screening test \cite{Ren2022_GlobalBCScreening}, while breast ultrasound (BUS) is widely used as a complementary exam because it is safe, real-time, portable, and relatively low-cost. Unlike mammography, BUS uses no ionizing radiation and is particularly helpful in dense breasts \cite{Vegunta2021_DenseBreastSupplemental}. Despite these advantages, BUS interpretation is challenging: speckle noise, acoustic shadowing, operator dependence, and low contrast often obscure lesion boundaries, making manual assessment variable and time-intensive. In clinical workflows, segmentation delineates lesion extent so that size, shape, and margins can be measured consistently and passed to downstream computer-aided diagnosis (CAD) systems. Automatic BUS segmentation has become a central task in computer-aided analysis, and several public benchmarks now support systematic evaluation on challenging lesions \cite{Zhang2022_BUSIS}.

Over the past decade, encoder--decoder architectures (e.g., U\hbox{-}Net variants) have become the standard backbone for BUS segmentation \cite{Ronneberger2015UNet,Isensee2021nnUNet}. Refinements such as attention, multi\hbox{-}scale aggregation and feature pyramids, dilated convolutions and larger receptive fields, residual and dense connections, and stronger decoders further improve the base design \cite{Lin2017FPN,Chen2018DeepLabV3Plus,He2016ResNet}. Beyond such generic upgrades, task\hbox{-}specific BUS methods target small or low\hbox{-}contrast lesions, speckle noise, and spiculated margins, showing gains on difficult cases \cite{Shareef2022_ESTAN_Healthcare}. Nevertheless, generalization across scanners, institutions, and patient demographics remains challenging when training data are narrow in scope. In parallel, Transformer\hbox{-}based and hybrid CNN--Transformer segmenters have matured, yet they still degrade under distribution shift without broader, multi\hbox{-}source supervision \cite{Chen2021_TransUNet}.

Prompt-based segmentation offers a complementary direction. Point or box prompts can steer masks with minimal interaction, and text-conditioned variants enable semantic guidance \cite{Kirillov2023SAM,Ma2024SAMMedical}. In BUS, text-guided approaches have begun to show promise by conditioning on short descriptors or standardized terms, improving delineation when such information is available \cite{MallinaShareef2025_XBusNet}. A practical limitation, however, is that many BUS datasets lack reliable prompts or metadata (e.g., BI-RADS descriptors) or provide them inconsistently. As a result, training is often restricted to the few multimodal datasets \cite{MallinaShareef2025_XBusNet,Zhang2022_BUSIS} or to small single-center cohorts, leaving larger image-only resources underutilized. This motivates methods that exploit prompts when present and remain robust when absent, so that mixed sources can be used without discarding data.

We address this constraint with a mixed-supervision framework that treats “no text” as a first-class state and integrates prompts at two complementary levels. Concretely, our model couples (i) a global pathway that leverages image-level context and (ii)  a local pathway that conditions mid- and low-level features. When text is present, short descriptors guide both pathways; when text is missing, nullable prompts—learnable null embeddings with presence masks—let the network fall back to image-only evidence without unstable imputations. This design enables training across heterogeneous datasets that differ in prompt availability while preserving strong image-only performance.  The most related work is ARSeg \cite{wang2025towards}, which tackles incomplete textual prompts in medical referring segmentation via attribute-specific cross-modal interactions and dual attribute losses. In contrast, we address absent prompts at the dataset level and train across mixed sources using learnable nullable embeddings, allowing effective use of large image-only cohorts.

\noindent\textbf{Our contributions are as follows.}
(i) We propose NullBUS, a dual-path prompt-aware BUS segmentation network that combines global image-level and local feature-level conditioning.
(ii) We introduce nullable prompts with learnable null embeddings and presence masks, enabling joint training on images with and without text.
(iii) We establish a mixed-dataset evaluation on three public BUS datasets and show that NullBUS improves IoU/Dice and reduces FNR compared with strong image-only and prompt-based baselines.

\section{Method}

\subsection{Overview}
\begin{figure*}[!t]
  \centering
  \includegraphics[width=0.75\textwidth]{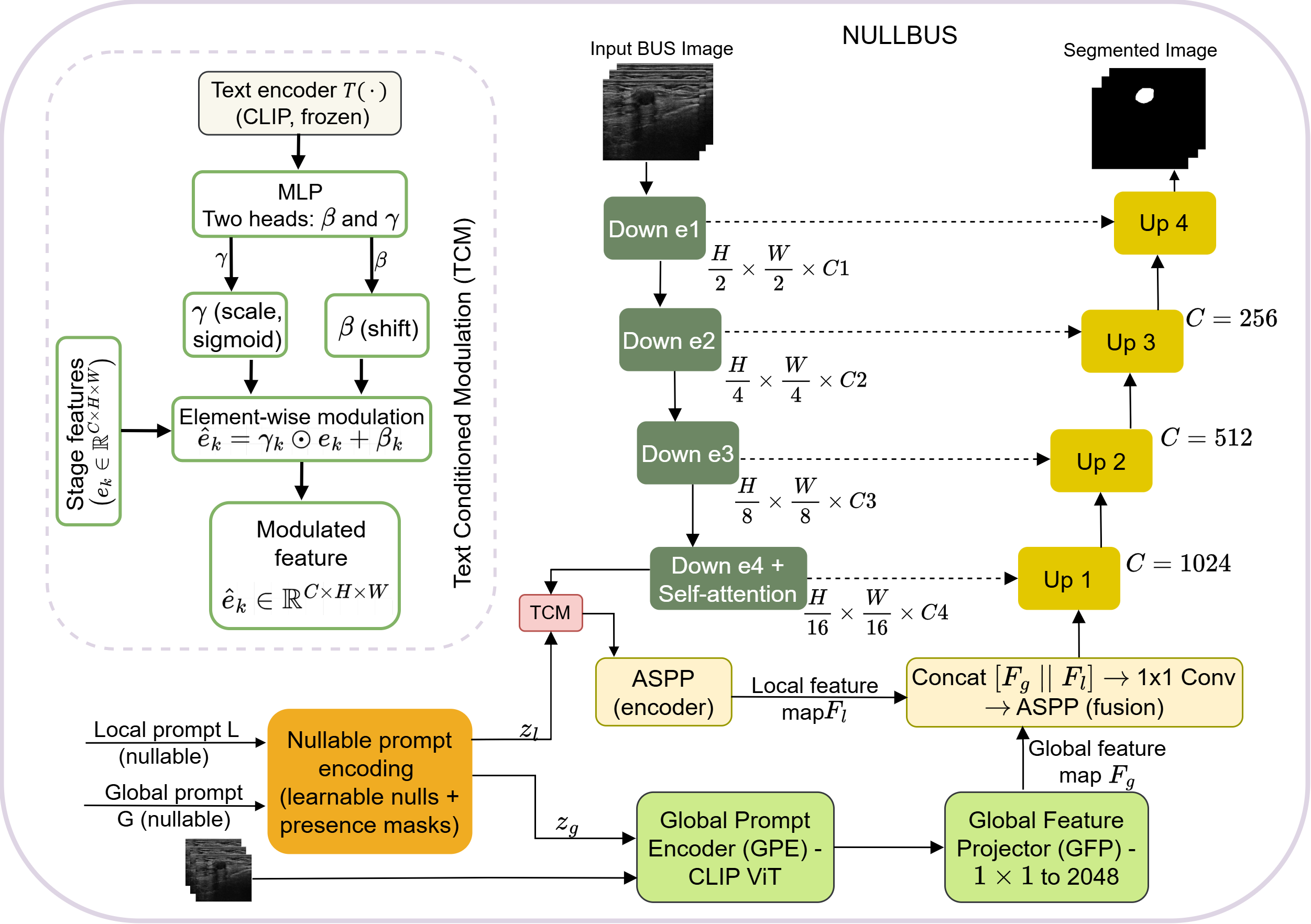}
  \caption{NullBUS architecture: dual-path BUS segmentation with nullable global and local prompts.}
  \label{fig:method}
\end{figure*}

We propose NullBUS, a prompt-optional segmentation framework for breast ultrasound. The model comprises two visual paths and a U-shaped decoder (Fig.~\ref{fig:method}). The global path provides image-level, text-conditioned context using a frozen CLIP ViT, implemented as a Global Prompt Encoder (GPE), followed by a Global Feature Projector (GFP) that produces a 2048-channel feature map. The local path extracts spatial detail with a ResNet-50 encoder; at its deepest stages, Text-Conditioned Modulation (TCM) injects prompt information and an ASPP bottleneck enlarges the receptive field. When text is missing at either level, nullable prompts—learnable null embeddings with presence masks—enable a reliable fallback to image-only evidence. Global and local features are fused at the bottleneck and decoded by four UpFusion stages to produce a \(352{\times}352\) mask, and a Prediction Head (1{\(\times\)}1 convolution) outputs the logits.

\subsection{Nullable prompt encoding}
Let \(T:\mathcal{S}\to\mathbb{R}^{d}\) be the text encoder and let \(G,L\in\mathcal{S}\cup\{\varnothing\}\) denote the global and local prompts.
We introduce learnable null embeddings \(z_{\varnothing}^{g}, z_{\varnothing}^{l}\in\mathbb{R}^{d}\) and binary presence masks
\(m_g=\Ind[\,G\neq\varnothing\,]\), \(m_l=\Ind[\,L\neq\varnothing\,]\).
During training, prompt dropout with rate \(p\in[0,1]\) samples \(d_g,d_l\sim\mathrm{Bernoulli}(1-p)\) and sets
\(\alpha_g = m_g\,d_g\), \(\alpha_l = m_l\,d_l\).
The embeddings passed to the network are
\[
z_g=\alpha_g\,T(G)+(1-\alpha_g)\,z_{\varnothing}^{g},\quad
z_l=\alpha_l\,T(L)+(1-\alpha_l)\,z_{\varnothing}^{l}. \tag{1}
\]
At inference, \(d_g=d_l=1\), so each path uses \(T(\cdot)\) when a prompt is present and its learned null otherwise.
Nulls are initialized from a neutral token and optimized jointly with model parameters.

\subsubsection*{How conditioning is applied}
In the global path, a light conditional blend modulates token features \(A\) using two MLP heads \(\gamma(\cdot),\beta(\cdot)\):
\[
\widetilde{A}=\gamma(z_g)\odot A+\beta(z_g), \tag{2}
\]
after which tokens are spatialized and projected by the GFP. In the local path, TCM injects \(z_l\) at deep stages \(e_k\) via channel-wise scale and shift,
\[
\widehat{e}_k=\gamma_k(z_l)\odot e_k+\beta_k(z_l),\qquad k\in\{4,5\}. \tag{3}
\]
When a prompt is absent, the same transforms operate with the learned null rather than zero-fill, providing stable text-free conditioning.

\subsubsection*{Unified handling of prompt regimes}
Equation~(1) yields a single forward pass that covers all regimes: (i) both prompts, \(z_g{=}T(G)\), \(z_l{=}T(L)\); (ii) global only, \(z_l{=}z_{\varnothing}^{l}\); (iii) local only, \(z_g{=}z_{\varnothing}^{g}\); (iv) neither, both paths use their learned nulls. With dropout, \(\mathbb{E}[z_g]=(1-p)m_g\,T(G)+\big(1-(1-p)m_g\big)z_{\varnothing}^{g}\) (and analogously for \(z_l\)), which regularizes the model under partial or inconsistent text availability.

\subsection{Global path}
GPE (frozen CLIP ViT-B/16) extracts intermediate token features that are modulated by \(z_g\) via Eq.~(2). Tokens are reshaped to a grid and upsampled to a spatial map; GFP applies a 1{\(\times\)}1 projection with GroupNorm to produce a 2048-channel map \(F_g\) aligned to the local bottleneck resolution.

\subsection{Local path}
ResNet-50 yields stage features \(\{e_1,\dots,e_5\}\). Around the deepest blocks, two self-attention layers enlarge context. TCM applies Eq.~(3) at the deep stages to incorporate \(z_l\). An ASPP module with rates \((1,6,12,18)\) forms the local bottleneck feature \(F_l\). Standard skip connections from \(e_1\!\to\!e_4\) feed the decoder.

\subsection{Fusion and decoding}
The bottleneck features are concatenated and refined:
\[
\mathrm{Fuse} = \mathrm{ASPP}\!\big(\mathrm{Conv}_{1{\times}1}([F_g \,\|\, F_l])\big).
\]
The decoder comprises four UpFusion stages that upsample, fuse with the corresponding skip, apply squeeze–excitation, and refine with a depthwise–pointwise residual block. A final \(2{\times}\) upsampler and a 1{\(\times\)}1 Prediction Head produce the logit map, which is bilinearly upsampled to \(352{\times}352\).

\subsection{Implementation and Optimization}
CLIP ViT-B/16 in GPE is kept frozen, while the local path uses a trainable ResNet-50 backbone. GroupNorm is used in all convolutional blocks. GFP outputs 2{,}048 channels, and the decoder predicts masks at a fixed resolution of \(352{\times}352\).
We minimize Dice loss on probabilities $\hat p = \sigma(\hat y)$ and mask $y$:
$\mathcal{L}_{\mathrm{Dice}} = 1 - \frac{2\langle \hat p, y\rangle + \varepsilon}{\lVert \hat p\rVert_1 + \lVert y\rVert_1 + \varepsilon}.$

\FloatBarrier
\begin{table}[H]
\centering
\caption{Public BUS datasets used to form the unified pool. Only images with pixel-wise lesion masks are included.}
\label{tab:nullbus-datasets}
\setlength{\tabcolsep}{5pt}
\renewcommand{\arraystretch}{1.05}
\resizebox{\columnwidth}{!}{%
\begin{tabular}{l|cccc}
\hline
\textbf{Dataset} & \textbf{Size} & \textbf{Distribution} & \textbf{Metadata} \\
\hline
BLU \cite{pawlowska2024curated}    & 252      & B = 154, M = 98 & Yes \\
BUSI \cite{ALDHABYANI2020104863}    & 630      & B = 431, M = 209         & No \\
BUSBRA \cite{gomez2024bus}  & 1{,}875 & B = 1268, M = 607              & Yes \\
\hline
\textbf{Total} & \textbf{2{,}757} &B = 1853, M = 914 & &  \\
\hline
\end{tabular}%
}
\vspace{-2mm}
\end{table}
\section{Experimental Results}
\subsection{Dataset and Setup}
We evaluate on a unified pool constructed from three public breast–ultrasound datasets: BLU \cite{pawlowska2024curated}  , BUSI \cite{ALDHABYANI2020104863}, and BUSBRA \cite{gomez2024bus} (Table~\ref{tab:nullbus-datasets}). Only images with pixel-wise masks are retained. BLU and BUSBRA provide non-uniform metadata (BI–RADS, pathology, limited patient history), while BUSI includes masks but no prompts. This mix reflects practical variability in prompt availability and motivates the nullable–prompt setting.

\noindent\textbf{Implementation details.}
All images are grayscale-normalized and resized to \(352{\times}352\). Training uses 5-fold image-level cross-validation with class-stratified folds. Training and inference are implemented in PyTorch, and all methods share the same preprocessing and data splits for fair comparison.

For prompt handling, NullBUS trains on the full pool: images with metadata contribute text embeddings, and images without metadata are automatically handled by the learned null embeddings in both the global and local branches.

\noindent\textbf{Evaluation metrics.}
We report Dice, IoU, and the error rates FPR and FNR, all computed from pixel-wise confusion counts (TP, FP, FN, TN) at a fixed threshold of 0.5. Dice and IoU are used as the primary segmentation metrics, while FPR and FNR quantify background and miss errors, respectively. Performance is measured under 5-fold image-level cross-validation. Table~\ref{tab:panseg-results} summarizes the 5-fold results.

\noindent\textbf{Comparison methods}
We compare NullBUS to five recent baselines spanning three categories:
(i) CNN baselines: U-ResNet and SANU-Net \cite{zhang2025sanu}, which rely on convolutional priors (with attention modules in SANU-Net) and serve as strong image-only segmentors. 
(ii) CNN–Transformer hybrid: FET-UNet \cite{zhang2025fet}, which augments a UNet-style CNN with Transformer blocks to capture long-range context.
(iii) Prompt-based: BUSSAM \cite{tu2024ultrasound} (vision prompts such as points/boxes) and ARSeg \cite{wang2025towards} (text prompts).

All baselines are retrained on our unified pool with identical preprocessing and supervision. For prompt-based models, prompts are supplied only when available in the source metadata; we do not synthesize missing prompts. In contrast, NullBUS uses nullable prompts and therefore trains on the entire pool and runs with or without text at inference, enabling a direct robustness comparison under mixed prompt availability.

\FloatBarrier
\begin{table}[H]
\centering
\caption{Overall Performance Comparison}
\label{tab:comparison-results}
\setlength{\tabcolsep}{6pt}
\renewcommand{\arraystretch}{1.05}
\begin{tabular}{l|cccc}
\hline
\textbf{Model} & \textbf{IoU} & \textbf{Dice} & \textbf{FPR} & \textbf{FNR} \\
\hline
ARSeg \cite{wang2025towards}          & 0.6225 & 0.7332 & 0.0211 & 0.2111 \\
FET-UNet \cite{zhang2025fet}       & 0.7303 & 0.8172 & 0.0123 & 0.1828 \\
SANU-Net \cite{zhang2025sanu}       & 0.7405 & 0.8264 & \textbf{0.0096} & 0.1745 \\
U-ResNet       & 0.7649 & 0.8303 & 0.0097 & 0.1557 \\
BUSSAM \cite{tu2024ultrasound}         & 0.7566 & 0.8416 & 0.0145 & 0.1292 \\
\textbf{NullBUS (ours)}  & \textbf{0.8568} & \textbf{0.9103} & 0.1847 & \textbf{0.0698} \\
\hline
\end{tabular}
\vspace{-2mm}
\end{table}
\subsection{Overall Performance}
\begin{figure}[!t]
  \centering
  \includegraphics[width=\columnwidth]{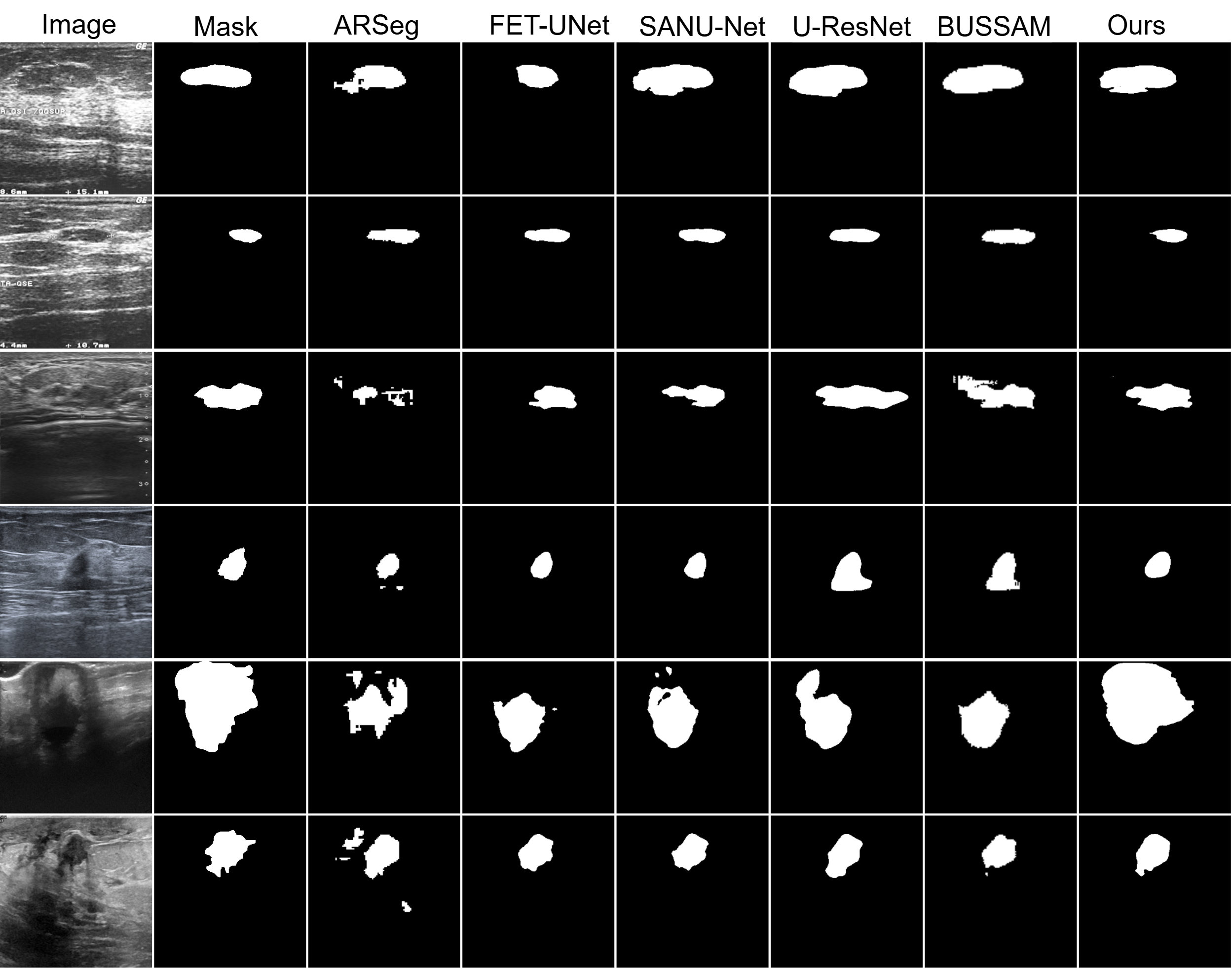}
  \caption{Qualitative comparison.}
  \label{fig:comparison}
\end{figure}

\noindent\textbf{Overall performance.}
Across the unified evaluation pool, NullBUS attains the highest overlap scores among all compared methods (Table~\ref{tab:comparison-results}). In particular, it achieves an IoU of 0.8568 and a Dice of 0.9103, surpassing the strongest baseline by approximately \(+9.2\) IoU points and \(+6.9\) Dice points. NullBUS also yields the lowest false–negative rate (FNR \(=\) 0.0698), indicating fewer missed lesions. Compared with the image-only baselines, this reduction in FNR is accompanied by a higher false–positive rate (FPR \(\approx 0.1847\) versus roughly \(0.01\text{--}0.02\)), reflecting a preference for sensitivity over background specificity. Across the five folds (Table~\ref{tab:panseg-results}), IoU ranges from \(0.8462\) to \(0.8664\) and Dice from \(0.9019\) to \(0.9173\), indicating stable generalization across splits. Qualitative results in Fig.~\ref{fig:comparison} show crisper boundaries and more complete coverage of small, low–contrast lesions, especially in challenging acoustic conditions, while avoiding over-smoothing that can remove fine spiculations.

\FloatBarrier
\begin{table}[H]
\centering
\caption{5-Fold Cross-Validation Results}
\label{tab:panseg-results}
\setlength{\tabcolsep}{5pt}
\renewcommand{\arraystretch}{1.05}
\begin{tabular}{c|cccc}
\hline
\textbf{Fold} & \textbf{IoU} & \textbf{Dice} & \textbf{FPR} & \textbf{FNR} \\
\hline
0 & 0.8664 & 0.9173 & 0.1858 & 0.0731 \\
1 & 0.8614 & 0.9138 & 0.1865 & 0.0840 \\
2 & 0.8597 & 0.9129 & 0.1847 & 0.0741 \\
3 & 0.8462 & 0.9019 & 0.1833 & 0.0585 \\
4 & 0.8503 & 0.9054 & 0.1835 & 0.0594 \\
\hline
Mean & 0.8568 & 0.9103 & 0.1847 & 0.0698 \\
\hline
\end{tabular}
\vspace{-2mm}
\end{table}

\subsection{Ablation studies}
We ablate the main NullBUS components in Table~\ref{tab:nullbus-ablation}.
Removing the global path (GPE+GFP) causes a modest drop of about 1.8 IoU and 1.5 Dice, so image-level context helps but is not the main driver.
Removing the local high-resolution path causes a large decline of about 20.3 IoU and 16.0 Dice and increases FNR, showing that fine-scale features are critical for boundaries.
A zero-text configuration, trained and evaluated without semantic prompts (both paths use learned nulls), collapses to IoU 0.363 with FNR 0.461, indicating that prompts provide essential guidance rather than mild regularization.
The full model achieves the best tradeoff, with the highest IoU and Dice and the lowest FNR among variants.

\FloatBarrier
\begin{table}[H]
\centering
\caption{Ablation on NullBUS components.}
\label{tab:nullbus-ablation}
\setlength{\tabcolsep}{5pt}
\renewcommand{\arraystretch}{1.05}
\resizebox{\columnwidth}{!}{%
\begin{tabular}{l|cccc}
\hline
\textbf{Experiment} & \textbf{IoU} & \textbf{Dice} & \textbf{FPR} & \textbf{FNR} \\
\hline
Zero Text-Guidance    & 0.3628 & 0.4385 & 0.1785 & 0.4613 \\
Zero Local Features   & 0.6543 & 0.7503 & \textbf{0.1744} & 0.0987 \\
Zero Global Features  & 0.8389 & 0.8956 & 0.1847 & 0.0897 \\
\textbf{NullBus} & \textbf{0.8568} & \textbf{0.9103} & 0.1847 & \textbf{0.0698} \\
\hline
\end{tabular}%
}
\vspace{-2mm}
\end{table}

\section{Conclusion}
We presented NullBUS, a multimodal mixed-supervision framework for breast ultrasound segmentation that treats missing text as a first-class state via nullable prompts. By coupling a global vision–language path with a local high-resolution path and decoding with a U-shaped head, the model achieves state-of-the-art overlap with stable 5-fold performance; ablations confirm complementary contributions from global context, local modulation, and nullable prompts. Limitations include public datasets with non-uniform metadata, class imbalance, and limited clinical diversity, and future work will pursue multicenter validation, automatic prompt extraction, and extensions to other modalities.

\bibliographystyle{IEEEbib}
\bibliography{refs}

@inproceedings{Ronneberger2015UNet,
  author    = {Olaf Ronneberger and Philipp Fischer and Thomas Brox},
  title     = {U-Net: Convolutional Networks for Biomedical Image Segmentation},
  booktitle = {MICCAI},
  year      = {2015},
  pages     = {234--241}
}

@inproceedings{wang2025towards,
  title={Towards Robust Medical Image Referring Segmentation with Incomplete Textual Prompts},
  author={Wang, Qijie and Lin, Xian and Yan, Zengqiang},
  booktitle={International Conference on Medical Image Computing and Computer-Assisted Intervention},
  pages={636--646},
  year={2025},
  organization={Springer}
}

@article{Isensee2021nnUNet,
  author  = {Isensee, Fabian and Jaeger, Paul F. and Kohl, Simon A. A. and Petersen, Jens and Maier-Hein, Klaus H.},
  title   = {nnU-Net: a self-configuring method for deep learning-based biomedical image segmentation},
  journal = {Nature Methods},
  year    = {2021},
  volume  = {18},
  number  = {2},
  pages   = {203--211}
}

@inproceedings{Lin2017FPN,
  title     = {Feature Pyramid Networks for Object Detection},
  author    = {Lin, Tsung-Yi and Doll{\'a}r, Piotr and Girshick, Ross and He, Kaiming and Hariharan, Bharath and Belongie, Serge},
  booktitle = {Proceedings of the IEEE Conference on Computer Vision and Pattern Recognition (CVPR)},
  pages     = {2117--2125},
  year      = {2017},
  doi       = {10.1109/CVPR.2017.106}
}

@inproceedings{Chen2018DeepLabV3Plus,
  author    = {Chen, Liang-Chieh and Zhu, Yukun and Papandreou, George and Schroff, Florian and Adam, Hartwig},
  title     = {Encoder-Decoder with Atrous Separable Convolution for Semantic Image Segmentation},
  booktitle = {Proceedings of the European Conference on Computer Vision (ECCV)},
  year      = {2018},
  pages     = {801--818}
}

@inproceedings{He2016ResNet,
  author    = {He, Kaiming and Zhang, Xiangyu and Ren, Shaoqing and Sun, Jian},
  title     = {Deep Residual Learning for Image Recognition},
  booktitle = {Proceedings of the IEEE Conference on Computer Vision and Pattern Recognition (CVPR)},
  year      = {2016},
  pages     = {770--778},
  doi       = {10.1109/CVPR.2016.90},
  url       = {https://doi.org/10.1109/CVPR.2016.90}
}

@article{Kirillov2023SAM,
  author  = {Kirillov, Alexander and Mintun, Eric and Ravi, Nikhila and Mao, Hanzi and Rolland, Chloe and Gustafson, Laura and Xiao, Tete and Whitehead, Spencer and Berg, Alex and Lo, Wan-Yen and Doll{\'a}r, Piotr and Girshick, Ross},
  title   = {Segment Anything},
  journal = {arXiv:2304.02643},
  year    = {2023}
}

@article{Ma2024SAMMedical,
  author  = {Jun Ma and others},
  title   = {Segment Anything in Medical Images},
  journal = {Nature Communications},
  year    = {2024},
  volume  = {15},
  number  = {1},
  pages   = {1022}
}

@article{zhang2025fet,
  title={FET-UNet: Merging CNN and transformer architectures for superior breast ultrasound image segmentation},
  author={Zhang, Huaikun and Lian, Jing and Ma, Yide},
  journal={Physica Medica},
  volume={133},
  pages={104969},
  year={2025},
  publisher={Elsevier}
}

@article{MallinaShareef2025_XBusNet,
  title   = {XBusNet: Text-Guided Breast Ultrasound Segmentation via Multimodal Vision-Language Learning},
  author  = {Mallina, Raja and Shareef, Bryar},
  journal = {Diagnostics},
  year    = {2025},
  volume  = {15},
  number  = {22},
  pages   = {2849},
  doi     = {10.3390/diagnostics15222849},
  url     = {https://doi.org/10.3390/diagnostics15222849}
}

@inproceedings{zhang2025sanu,
  title={SANU-Net: An Improved NU-Net Breast Ultrasound Image Segmentation Method Combined with Attention Mechanism},
  author={Zhang, Chuxuan and Ren, Jieyu and Zhang, Yuehan and Dai, Yu and Zhou, Lu},
  booktitle={2025 44th Chinese Control Conference (CCC)},
  pages={8092--8096},
  year={2025},
  organization={IEEE}
}

@article{Shareef2022_ESTAN_Healthcare,
  title   = {ESTAN: Enhanced Small Tumor-Aware Network for Breast Ultrasound Image Segmentation},
  author  = {Shareef, Bryar and Vakanski, Aleksandar and Freer, Phoebe E. and Xian, Min},
  journal = {Healthcare},
  volume  = {10},
  number  = {11},
  pages   = {2262},
  year    = {2022},
  doi     = {10.3390/healthcare10112262}
}

@article{tu2024ultrasound,
  title={Ultrasound sam adapter: Adapting sam for breast lesion segmentation in ultrasound images},
  author={Tu, Zhengzheng and Gu, Le and Wang, Xixi and Jiang, Bo},
  journal={arXiv preprint arXiv:2404.14837},
  year={2024}
}

@article{ALDHABYANI2020104863,
title = {Dataset of breast ultrasound images},
journal = {Data in Brief},
volume = {28},
pages = {104863},
year = {2020},
issn = {2352-3409},
doi = {https://doi.org/10.1016/j.dib.2019.104863},
url = {https://www.sciencedirect.com/science/article/pii/S2352340919312181},
author = {Walid Al-Dhabyani and Mohammed Gomaa and Hussien Khaled and Aly Fahmy},
}

@article{pawlowska2024curated,
  title={Curated benchmark dataset for ultrasound based breast lesion analysis},
  author={Paw{\l}owska, Anna and {\'C}wierz-Pie{\'n}kowska, Anna and Domalik, Agnieszka and Jagu{\'s}, Dominika and Kasprzak, Piotr and Matkowski, Rafa{\l} and Fura, {\L}ukasz and Nowicki, Andrzej and {\.Z}o{\l}ek, Norbert},
  journal={Scientific Data},
  volume={11},
  number={1},
  pages={148},
  year={2024},
  publisher={Nature Publishing Group UK London}
}

@techreport{ACS_Breast_Common_2025,
  title        = {Cancer Facts \& Figures 2025},
    author = {American Cancer Society},
    institution = {American Cancer Society},
  year         = {2025},
  url          = {https://www.cancer.org/content/dam/cancer-org/research/cancer-facts-and-statistics/annual-cancer-facts-and-figures/2025/2025-cancer-facts-and-figures-acs.pdf}
}

@article{Ren2022_GlobalBCScreening,
  title   = {Global guidelines for breast cancer screening: A systematic review},
  author  = {Ren, Wenhui and Chen, Mingyang and Qiao, Youlin and Zhao, Fanghui},
  journal = {Breast},
  year    = {2022},
  volume  = {64},
  pages   = {85--99},
  doi     = {10.1016/j.breast.2022.04.003},
  url     = {https://doi.org/10.1016/j.breast.2022.04.003}
}

@article{Vegunta2021_DenseBreastSupplemental,
  title   = {Supplemental Cancer Screening for Women With Dense Breasts: Guidance for Health Care Professionals},
  author  = {Vegunta, Suneela and Kling, Juliana M. and Patel, Bhavika K.},
  journal = {Mayo Clinic Proceedings},
  year    = {2021},
  volume  = {96},
  number  = {11},
  pages   = {2891--2904},
  doi     = {10.1016/j.mayocp.2021.06.001},
  url     = {https://doi.org/10.1016/j.mayocp.2021.06.001}
}

@article{Chen2021_TransUNet,
  title   = {TransUNet: Transformers Make Strong Encoders for Medical Image Segmentation},
  author  = {Chen, Jieneng and others},
  journal = {arXiv:2102.04306},
  year    = {2021},
  url     = {https://arxiv.org/abs/2102.04306}
}

@article{gomez2024bus,
  title={BUS-BRA: a breast ultrasound dataset for assessing computer-aided diagnosis systems},
  author={G{\'o}mez-Flores, Wilfrido and Gregorio-Calas, Maria Julia and Coelho de Albuquerque Pereira, Wagner},
  journal={Medical Physics},
  volume={51},
  number={4},
  pages={3110--3123},
  year={2024},
  publisher={Wiley Online Library}
}

@article{Zhang2022_BUSIS,
  title   = {BUSIS: A Benchmark for Breast Ultrasound Image Segmentation},
  author  = {Zhang, Yifeng and Xian, Ming and Cheng, Hsiang-Chih and Shareef, Bryar and Ding, Jing and Xu, Fang and Huang, Kun and Zhang, Bo and Ning, Changhong and Wang, Yongyi},
  journal = {Healthcare},
  year    = {2022},
  volume  = {10},
  number  = {4},
  pages   = {729},
  doi     = {10.3390/healthcare10040729},
  url     = {https://doi.org/10.3390/healthcare10040729}
}

\end{document}